\documentclass[11pt,a4paper]{article}
\usepackage[hyperref]{emnlp2020}
\usepackage{times}
\usepackage{latexsym}

\usepackage{microtype}

\aclfinalcopy 

\usepackage{todonotes}

\usepackage{enumitem}
\usepackage{booktabs}
\usepackage{multirow}
\usepackage{amsmath}
\usepackage{amsfonts}
\usepackage{fontawesome}
\usepackage[ruled,vlined]{algorithm2e}
\usepackage{arydshln}
\usepackage[procnames]{listings}
\usepackage{multicol}
\usepackage[title]{appendix}
\newcommand{\argmax}[1]{\underset{#1}{\operatorname{arg}\,\operatorname{max}}\;}

\title{Interpretable Entity Representations through Large-Scale Typing}

\author{Yasumasa Onoe \and Greg Durrett\\
  Department of Computer Science \\
  The University of Texas at Austin \\
  {\tt\{yasumasa, gdurrett\}@cs.utexas.edu}}

\date{}

\begin{document}
\maketitle


\begin{abstract}
In standard methodology for natural language processing, entities in text are typically embedded in dense vector spaces with pre-trained models. The embeddings produced this way are effective when fed into downstream models, but they require end-task fine-tuning and are fundamentally difficult to interpret. In this paper, we present an approach to creating entity representations that are human readable and achieve high performance on entity-related tasks out of the box. Our representations are vectors whose values correspond to posterior probabilities over fine-grained entity types, indicating the confidence of a typing model’s decision that the entity belongs to the corresponding type. We obtain these representations using a fine-grained entity typing model, trained either on supervised ultra-fine entity typing data \citep{Eunsol_Choi_18} or distantly-supervised examples from Wikipedia. On entity probing tasks involving recognizing entity identity, our embeddings used in parameter-free downstream models achieve competitive performance with ELMo- and BERT-based embeddings in trained models. We also show that it is possible to reduce the size of our type set in a learning-based way for particular domains. Finally, we show that these embeddings can be post-hoc modified through a small number of rules to incorporate domain knowledge and improve performance.
\end{abstract}

\section{Introduction}
In typical neural NLP systems, entities are embedded in the same space as other words either in context-independent \citep{Tomas_Mikolov_13, Jeffrey_Pennington_14} or in context-dependent ways \citep{Matthew_Peters_18, Jacob_Devlin_19}. Such approaches are powerful: pre-trained language models implicitly learn factual knowledge about those entities \citep{Fabio_Petroni_19,Adam_Roberts_20,Zhengbao_Jiang_20} and these representations can be grounded in structured and human-curated knowledge bases \citep{Robert_Logan_19,Yoav_Levine_19,Matthew_Peters_19,Zhengyan_Zhang_19,Nina_Poerner_19,Wenhan_Xiong_20,Ruize_Wang_20}. However, these embeddings do not \emph{explicitly} maintain representations of this knowledge, and dense entity representations are not directly interpretable. Knowledge probing tasks can be used to measure LMs' factual knowledge  \citep{Fabio_Petroni_19}, but designing the right probing task is another hard problem  \citep{Mingda_Chen_19,Nina_Poerner_19}, particularly if the probes are parameter-rich \cite{John_Hewitt_19_a}.


\begin{figure*}[!t]
    \centering
    \includegraphics[width=1.0\linewidth]{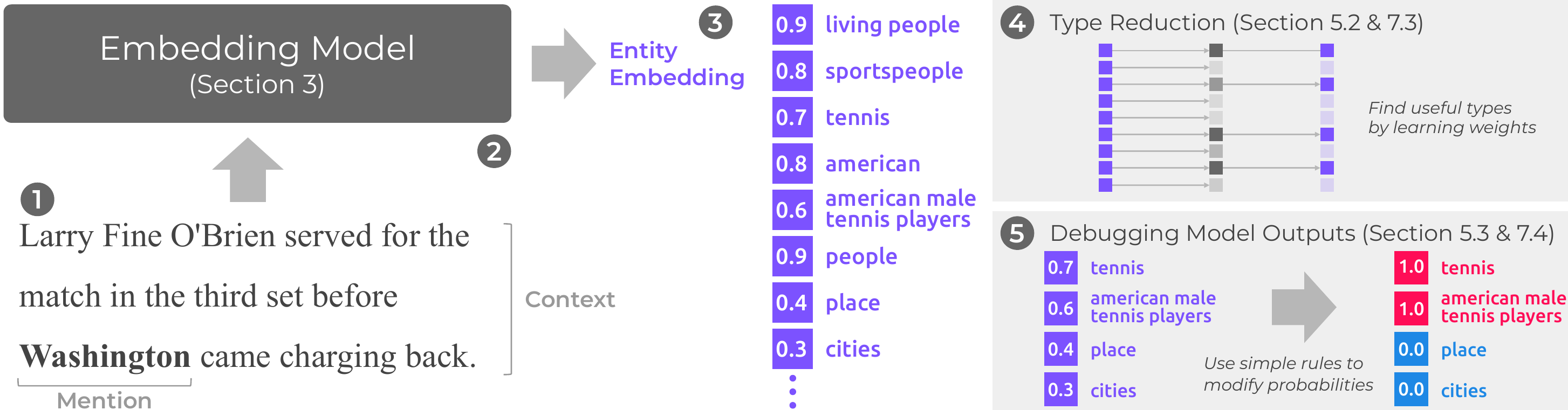}
    \caption{Interpretable entity representations. (1) A mention and its context are fed into (2) an embedding model. (3) An entity embedding vector consists of probabilities for corresponding types. (4) We can reduce the size of the type set for a particular downstream task in a learning-based way (Section~\ref{type-reduction}).  (5) We can also incorporate domain knowledge via rules to modify bad type probabilities and improve model performance (Section~\ref{Debuggability}).}
    \label{fig:approach}
\end{figure*}


In this work, we explore a set of interpretable entity representations that are simultaneously human and machine readable. The key idea of this approach is to use fine-grained entity typing models with large type inventories \citep{Xiao_Ling_12,Dan_Gillick_14,Eunsol_Choi_18, Yasumasa_Onoe_20}. Given an entity mention and context words, our typing model outputs a high-dimensional vector whose values are associated with predefined fine-grained entity types. Each value ranges between 0 and 1, corresponding to the confidence of the model's decision that the entity has the property given by the corresponding type. We use pre-trained Transformer-based entity typing models, trained either on a supervised entity typing dataset \cite{Eunsol_Choi_18} or on a distantly-supervised dataset derived from Wikipedia categories \cite{Yasumasa_Onoe_20}. The type vectors from these models, which contain tens of thousands of types, are then used as contextualized entity embeddings in downstream tasks.

Past work has shown that such type-driven representations are useful for entity linking \cite{Yasumasa_Onoe_20}; we improve the quality of these representations, broaden the scope of where they can be applied, and show techniques to extend and debug them by exploiting their interpretable nature. Dense representations of entities have similarly been applied to entity linking \citep{Ikuya_Yamada_16,Yotam_Eshel_17}, as well as relation extraction \citep{Baldini_Soares_19}, entity typing \citep{Jeffrey_Ling_20}, and question answering \citep{Thibault_Fevry_20}. Those approaches use millions of predefined entities, while our approach uses a much smaller number of types (10k or 60k). This makes it simultaneously more compact and also more flexible when generalizing to unknown entities.

We evaluate our embedding approach on benchmark tasks for entity representations. We use coreference arc prediction (CAP) and named entity disambiguation on CoNLL-YAGO, two tasks in the EntEval suite \citep{Mingda_Chen_19}, as well as entity linking on WikilinksNED \citep{Yotam_Eshel_17}, which covers broader entities and writing styles. We compare our approach against entity representations produced directly by pre-trained models. Our ``out-of-the-box'' entity representations combined with simple heuristics like dot product or cosine similarity can achieve competitive results on CAP without additional trainable parameters. On NED tasks, our approach outperforms all baselines by a substantial margin. We show that a much smaller type set can be distilled in a per-task fashion to yield similar performance. Finally, we show a proof-of-concept for how we can leverage the interpretability of our embeddings to ``debug'' downstream errors, a challenge for black-box models.

\section{Interpretable Entity Representations}

Our approach for producing entity representations is shown in Figure~\ref{fig:approach}. For an entity mention in context,\footnote{Our approach can also embed knowledge base entities, as discussed in Section~\ref{training}.} we compute a vector of probabilities, each of which reflects (independently) the probability of an entity exhibiting a particular type. Types are predefined concepts that could be derived from existing knowledge bases. We hypothesize that real world entities can be represented as a combination of those concepts if we have a large and varied enough concept inventory. This representation can be used as a dense vector since the values are still continuous numbers (though restricted between 0 and 1). It is interpretable like a discrete feature vector, since each dimension has been named with the corresponding entity type.

We define $s = (w_1, ..., w_N)$ to denote a sequence of context words, and $m = (w_i, ..., w_j)$ to denote an entity mention span in $s$. The input word sequence $s$ could be naturally co-occurring {\it context} words for the mention, or {\it descriptive} words such as might be found in a definition. The output variable is a vector $\mathbf{t} \in  [0,1]^{|\mathcal{T}|}$ whose values are probabilities corresponding to fine-grained entity types $\mathcal{T}$. These entity types are predefined and static, so their meanings are identical for all entities. Our goal here is to learn parameters $\theta$ of a function $f_\theta$ that maps the mention $m$ and its context $s$ to a vector $\mathbf{t}$, which capture salient features of the entity mention with the context.

We learn the parameters $\theta$ in a supervised manner. We use a labeled dataset $\mathcal{D} = \{(m, s, \mathbf{t}^*)^{(1)}, ... , (m, s, \mathbf{t}^*)^{(k)} \}$ to train an entity typing model. The gold labels $\mathbf{t}^*$ are obtained by manual annotation or distant-supervision techniques \cite{Mark_Craven_99,Mike_Mintz_09}. We select a predefined types $\mathcal{T}$ from modified Wikipedia categories, or we use an existing type set such as UFET \citep{Eunsol_Choi_18} (discussed in Section~\ref{training}).


\begin{figure}[!t]
    \centering
    \includegraphics[width=1.0\linewidth]{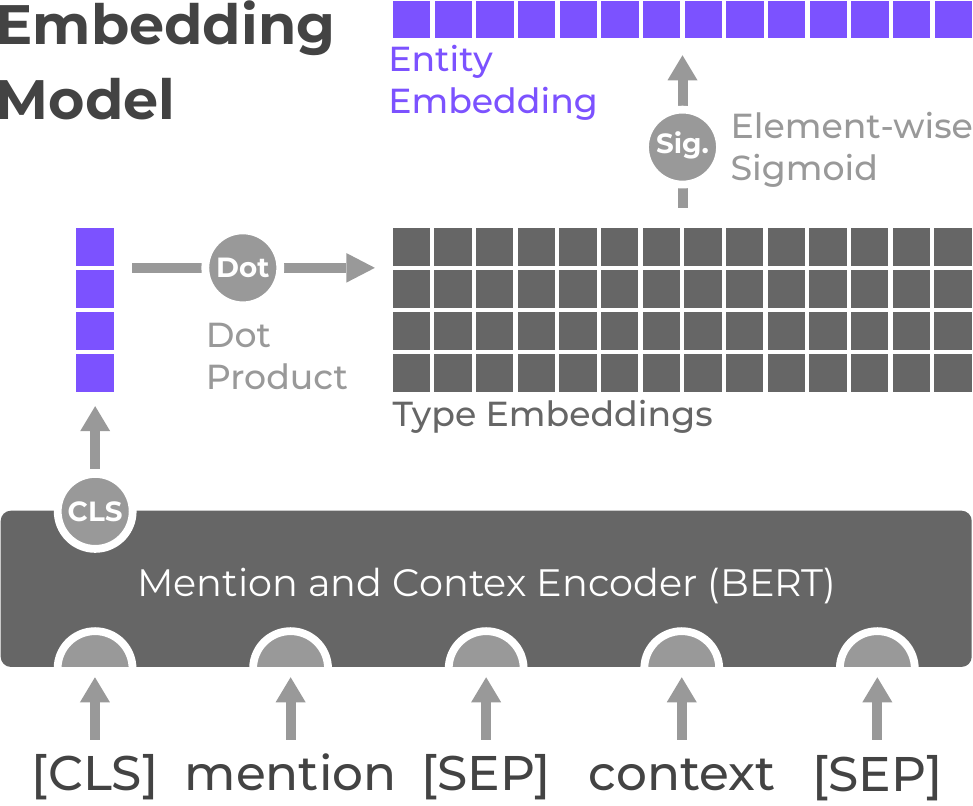}
    \caption{Embedding model architecture. We use BERT to embed the mention and context, then multiply by an output matrix and apply an elementwise sigmoid to compute posterior probabilities for each type. }
    \label{fig:model}
\end{figure}


We use the output vectors $\mathbf{t}$ as general purpose entity representations in downstream tasks, combined with off-the-shelf similarity measures like dot product and cosine similarity. These representations can also be customized in a per-task way. The number of entity types can be reduced by 90\% while maintaining similar performance (top right of Figure~\ref{fig:approach}), as discussed in Section~\ref{type-reduction}.

Another advantage of our interpretable embeddings is that they give us a hook to ``debug'' our downstream models. Debugging black-box models built on embeddings is typically challenging, but since our entity representations are directly interpretable, we can modify the output vectors $\mathbf{t}$ using our prior knowledge about entities (bottom right of Figure~\ref{fig:approach}). For example, we might know that in the financial domain {\it Wall Street} does not refer to a location. We show that simple rules based on prior knowledge can improve performance further (discussed in Section~\ref{Debuggability}); critically, this is done without having to annotate data in the target domain, giving system designers another technique for adapting these models.

\section{Embedding Model}\label{model}

Our model $f_\theta$ to produce these embeddings is shown in Figure~\ref{fig:model}: it takes as input the mention $m$ and its context $s$ and predicts probabilities for predefined entity types $\mathcal{T}$. This is a Transformer-based typing model following the BERT model presented in \citet{Yasumasa_Onoe_19}. First, a Transformer-based encoder \citep{Ashish_Vaswanir_17} maps the input variables, $m$ and $s$, to an intermediate vector representation. A type embedding layer then projects the intermediate representation to a vector whose dimensions correspond to the entity types $\mathcal{T}$. Finally, we apply a sigmoid function on each real-valued score in the vector to obtain the posterior probabilities that form our entity representation $\mathbf{t}$ (top of the figure).

\paragraph{Mention and Context Encoder} We use pre-trained BERT\footnote{We use BERT-large uncased (whole word masking) in our experiments. We experimented with RoBERTa \citep{Yinhan_Liu_19} but found it to work less well.} \citep{ Jacob_Devlin_19} for the mention and context encoder. This BERT-based encoder accepts as input a token sequence formatted as $\mathbf{x} =$ {\tt[CLS]} $m$  {\tt[SEP]} $s$ {\tt[SEP]}, where the mention $m$ and context $s$ are chunked into WordPiece tokens \citep{Yonghui_Wu_16}. We encode the whole sequence using BERT and use the hidden vector at the {\tt [CLS]} token as the mention and context representation: $\mathbf{h}_{\texttt{[CLS]}} = \textsc{BertEncoder}(\mathbf{x})$.

\paragraph{Type Embeddings} This output layer is a single linear layer whose parameter matrix can be viewed as a matrix of type embeddings $\mathbf{E} \in  \mathbb{R}^{|\mathcal{T}| \times d}$, where $d$ is the dimension of the mention and context representation $\mathbf{h}_{\texttt{[CLS]}}$. We obtain the output probabilities $\mathbf{t}$ by multiplying $\mathbf{E}$ by $\mathbf{h}_{\texttt{[CLS]}}$, followed by an element-wise sigmoid function: $\mathbf{t} = \sigma \left(\mathbf{E} \cdot \mathbf{h}_{\texttt{[CLS]}}\right)$.\footnote{Note that this makes our entities occupy a $d$-dimensional subspace in the type representation logit space (pre-sigmoid). A different model could be used to combat this low-rankness. Regardless, the explicit type space has advantages in terms of out-of-the-box functionality as well as interpretability.} Similar to previous work \citep{Eunsol_Choi_18, Yasumasa_Onoe_19}, we assume independence between all entity type in $\mathcal{T}$. 

One assumption in our approach is that the model's output probabilities are a meaningful measure of class membership. Past work \citep{Shrey_Desai_20} has observed that this is true for other models involving BERT variants.

\paragraph{Training}  Following \citet{Eunsol_Choi_18}, the loss is a sum of binary cross-entropy losses over all entity types $\mathcal{T}$ over all training examples $\mathcal{D}$. That is, we treat each type prediction for each example as an independent binary decision, with shared parameters in the BERT encoder.

\section{Training Data}\label{training}

To train our entity typing model, we need labeled examples consisting of $(m,s,\mathbf{t}^*)$ triples. Although there are labeled typing datasets such as UFET \citep{Eunsol_Choi_18}, getting large amounts of manually labeled data is expensive. Moreover, the UFET dataset contains instances of entities in context, so it is suitable for training models for \emph{contextual embeddings}, but it doesn't have examples of definitions for \emph{descriptive embeddings} (following the terminology of \citet{Mingda_Chen_19}). 

Therefore, we additionally use two distantly labeled entity typing datasets derived from Wikipedia. We leverage past work in using types derived from Wikipedia categories \cite[inter alia]{Suchanek07,Yasumasa_Onoe_20}, which contain type information and are widely annotated across Wikipedia articles. We select the appropriate dataset for each setting depending on task-specific requirements (see Section~\ref{experiments}). For all datasets, we compute entity typing macro F1 using development examples (1k) to check model convergence. 

\paragraph{Wiki-Context}\label{wiki-ctx} We collect a set of occurrences of typed entity mentions using hyperlinks in Wikipedia. Given a sentence with a hyperlink, we use the hyperlink as an entity mention $m$, the sentence as a context sentence $s$, and the Wiki categories of the destination page as the gold entity types $\mathbf{t}^*$. We use the preprocessing of \citet{Yasumasa_Onoe_20} to modify the type set: they introduce more general categories into the Wikipedia category set by splitting existing complex categories. Following their work, we filter the resulting set to keep the 60,000 most frequent types. Scraping Wikipedia yields 6M training examples that cover a wide range of entities and entity types. 

\paragraph{Wiki-Description}\label{wiki-desc}  Following a similar paradigm as for Wiki-Context, we create description-focused training examples from Wikipedia.\footnote{For tasks like entity linking, we could in principle just use gold type vectors for each entity, as in \citet{Yasumasa_Onoe_20}. However, the paradigm here matches that of \citet{Mingda_Chen_19}, and the descriptive entity embedding model we train can generalize to unseen descriptions at test time.} We use the same entity type set as the Wiki-Context dataset. We collect lead paragraphs from all Wikipedia pages and filter to keep examples that contain at least 1 entity type in the 60k entity types. We use the Wikipedia page title (usually boldfaced) in the lead paragraph as the entity mention $m$, and retain at most 100 words on either side to form the context $s$. The Wiki categories of the same page would be the gold entity types $\mathbf{t}^*$. We obtain 2M training examples after filtering. The size of entity type set is 60k.

\paragraph{UFET}\label{ufet}  This ultra-fine entity typing dataset is created by \citet{Eunsol_Choi_18}. This dataset consists of 6k manually annotated examples. The entity mention spans could be named entities, nominal expressions, and pronouns while Wiki-based datasets mostly provide named entity mention spans. We use 5.5k examples for training and 500 examples for validation. Note that because our goal in this work is downstream task performance, we deviate from the standard train/dev/test splits of 2k/2k/2k in favor of higher performance.


\section{Tailoring to a Task}\label{tailoring-to-a-task}
Our interpretable entity embeddings are designed for general-purpose uses and intended to work ``out-of-the-box". We first discuss two scenarios (tasks) and then describe two ways we can customize these representations for a downstream task: reducing the size of types and debugging model output using prior knowledge.

\subsection{Tasks}

\paragraph{Coreference Arc Prediction (CAP)}  This task focuses on resolving local coreference arcs. For each instance, two entity mention spans and their context are provided. The task is to predict if those two mention spans are coreferent or not, so this is a binary classification problem.\footnote{The mentions in this case are always drawn from the same or adjacent sentences, so constraints from saliency that would need to be incorporated in a full coreference system are less relevant in this setting.}

\paragraph{Named Entity Disambiguation (NED)} NED is the task of connecting entity mentions in text with real world entities in a knowledge base, including disambiguating between sometimes highly related candidates (e.g., the same movie produced in different years). We use the local resolution setting where each instance features a single entity mention span in the input text and several possible candidates. We consider the setting where descriptions for candidates entities are available (e.g., the first sentence of the Wikipedia article).


\subsection{Type Reduction}\label{type-reduction}
The type sets we consider in this work are very large, consisting of 10k or 60k types. Although larger type sets provides more precise entity representations, these may have redundant types or types which are unimportant for a particular domain. For both statistical and computational efficiency, we would like to compute the types useful for a downstream task in a data-driven way.

For all tasks we consider in this work, our model will depend chiefly on a function $\text{sim}(\mathbf{t}_1,\mathbf{t}_2)$ for two different type vectors. These type vectors are computed from mention and context pairs using the trained entity typing model $\mathbf{t} = f_\theta(m, s)$. In experiments, we will use either dot product or cosine similarity as our similarity function.



Our approach to compression involves learning a sparse trainable mask that restricts the set of types considered. We parameterize the dot product\footnote{$\text{sim}_{\text{dot}}(\mathbf{t}_1,\mathbf{t}_2) =  {\mathbf{t}_1}^\top \mathbf{W}\: \mathbf{t}_2$} and cosine similarity\footnote{$\text{sim}_{\text{cos}}(\mathbf{t}_1,\mathbf{t}_2) =\frac{\mathbf{t}_1^\top \mathbf{W}\: \mathbf{t}_2} {\sqrt{\mathbf{t}_1^\top \mathbf{W}\: \mathbf{t}_1} \sqrt{\mathbf{t}_2^\top \mathbf{W}\: \mathbf{t}_2}}$} operations with a weight matrix $\mathbf{W}$, a diagonal matrix $\text{diag}(w_1, w_2, ..., w_{|\mathcal{T}|})$ whose components correspond to the entity types in $\mathcal{T}$. The parameters $\mathbf{W}$ can be learned directly on downstream tasks (e.g., CAP and NED). Note that in the cosine scoring function, we clip these parameter values to be between 0 and 1. We train with the standard downstream task objective, but with an additional L$_1$ regularization term applied to $\mathbf{W}$ \citep{Robert_Tibshirani_94} to encourage the $\mathbf{W}$ values to be sparse. 

This approach naturally leads to around $20 - 35$\% sparsity in the vector $\text{diag}(w_1, w_2, ..., w_{|\mathcal{T}|})$ with settings of the regularization parameter we found effective. In practice, to achieve a higher level of sparsity, we further reduce the entity type set based on the magnitude of $\mathbf{W}$ (e.g., keep the 10\% of types with the highest values). Finally, we use the reduced entity types for further experiments on the target task.

\subsection{Debuggability}\label{Debuggability}

Our interpretable entity representations allow us to more easily understand when our models for downstream tasks make incorrect predictions, typically by mischaracterizing an ambiguous entity in context by assigning incorrect probabilities to the entity types. As an example from the CoNLL-YAGO NED dataset, we observe that our model gets confused if the mention span \emph{Spain} should refer to \emph{Women's national tennis team} or \emph{Men's national tennis team}. If we are trying to adapt to this scenario without using in-domain annotated data, a domain expert may nevertheless be able to articulate a rule to fix this error. Such a rule might be: whenever \emph{Fed Cup} (the international team competition in women's tennis) appears in the context, we assign 1 to a collection of relevant entity types such as \texttt{women's} and 0 to irrelevant types such as \texttt{davis cup teams} (the international team competition in men's tennis). Critically, because our representations have interpretable axes, we can more easily transform our entity representations and incorporate this kind of domain knowledge.

\section{Experimental Setup}\label{experiments}

We evaluate the ``out-of-the-box'' quality of our entity representations and baselines on two entity probing tasks as discussed in the previous section. 

\subsection{Datasets}

\paragraph{Coreference Arc Prediction (CAP)}

We use the English CAP dataset derived from PreCo \citep{Hong_Checn_18} by \citet{Mingda_Chen_19}. 
The creators of the dataset partition the data by cosine similarity of GloVe \cite{Jeffrey_Pennington_14} embeddings of mention spans and balance the number of positive and negative examples in each bucket, so that models do not solve the task by capturing surface features of entity mention spans. The original data split provides 8k examples for each of the training, development, and test sets.

\paragraph{Named Entity Disambiguation (NED)}
We use the standard English CoNLL-YAGO benchmark \citep{Johannes_Hoffart_11} preprocessed by \citet{Mingda_Chen_19}. For each entity mention, at most 30 candidate entities are selected using the CrossWikis dictionary \citep{Spitkovsky_12}. This dataset contains 18.5k training, 4.8k dev, and 4.5k test examples from newswire text, so the variety of entities and the writing styles are limited. For this reason, we create another NED dataset from WikilinksNED \citep{Yotam_Eshel_17}, which includes a wide range of entities and diverse writing styles from scraped English web text linking to Wikipedia. We limit the number of candidate entities to 3 for each instance, which still makes a challenging benchmark. We create 5k training, 1k dev, and 1k test examples and call this dataset WLNED. In both CoNLL-YAGO and WLNED, we form descriptions of candidate entities using the Wiki-Context data, but do not use any structural information from Wikipedia (hyperlinks, etc.).


\begin{figure}[!t]
    \centering
    \includegraphics[width=1.0\linewidth]{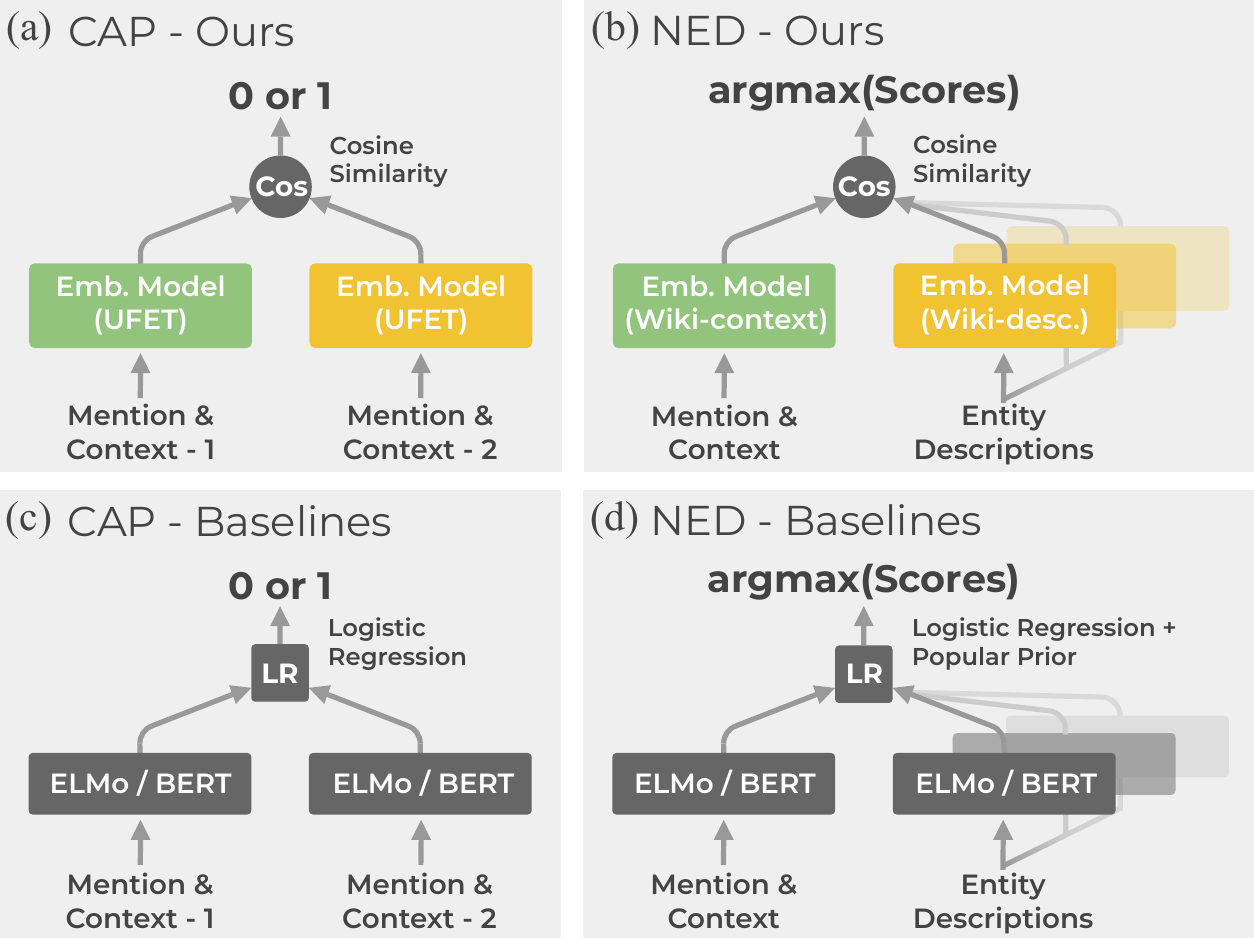}
    \caption{Overview of our downstream architectures. Our method simply computes cosine similarity and uses it as a score for each task, not introducing any new parameters. Our baselines use a trainable logistic regression layer over pre-trained embeddings to make classification decisions.} 
    \label{fig:downstream}
\end{figure}


\subsection{Baselines}

Figure~\ref{fig:downstream} schematically shows the use of our model compared to baselines, which we now describe.

\paragraph{Entity Embeddings}
We create entity representations of a mention span $m$ and a context $s$ using ELMo \citep{Matthew_Peters_18} and BERT \citep{Jacob_Devlin_19}. We largely follow the embedding procedure of \citet{Mingda_Chen_19}. Their downstream models use trainable weights to combine the vectors from the pre-trained model layers; we use this in the baselines as well except for the results in Table~\ref{tab:wlned} and for ELMo. Note that we do not fine tune the ELMo and BERT parameters of the baselines, since our focus is on \emph{general} entity representations that can work off-the-shelf rather than task specific entity representations. Our approach does not use task specific fine tuning either.
\begin{description}[align=left, font=\sc, topsep=2pt, labelsep=\fontdimen2\font, leftmargin=8pt]
\setlength\itemsep{-0.1em}
\item [ELMo]  \hspace{2pt} We run ELMo on the entire sentence $s$ and combine the three layer outputs using uniform weights. Then, we average contextualized vectors of the mention span $m$ to obtain the entity representation. 
\item [BERT Base] \hspace{2pt} We concatenate an entity mention $m$ and its context $s$ and feed it into BERT-base. We compute the weighted sum of the {\tt[CLS]} vectors\footnote{We tried pooling span representations like for ELMo and saw similar results.} from all 13 layers  and use it as an entity representation. 
\item [BERT Large] \hspace{2pt} Similar to the BERT-base baseline, we feed an entity mention $m$ and its context $s$ into BERT-large and average the {\tt[CLS]} vectors from all 25 layers.  
\item [KnowBert-W+W] \hspace{2pt} KnowBert is built on top of BERT-base by adding an internal entity linker. We use KnowBert-W+W, which has been trained on Wikipedia and WordNet \citep{Christiane_Fellbaum_98} as an \emph{embedding model} that incorporates external information; note that we are \emph{not} using this as an entity linking system, even for NED. Similar to other BERT baselines, we feed a mention span $m$ and context $s$, and we use the weighted sum of the {\tt[CLS]} vectors from all 15 layers.
\end{description}


%

\renewcommand{\arraystretch}{1}
\begin{table}[t]
	\centering
	\small
	\setlength{\tabcolsep}{4pt}
	\begin{tabular}{l c}
		\toprule
		\multicolumn{1}{l}{Model} & {Test Acc.}\\
		\midrule
		{\sc GloVe} \citep{Mingda_Chen_19} & 71.9   \\
		{\sc ELMo} \citep{Mingda_Chen_19} & 80.2   \\
		{\sc BERT Base} $\rightarrow$ LR \citep{Mingda_Chen_19} & 80.6  \\
		{\sc BERT Large} $\rightarrow$ LR \citep{Mingda_Chen_19} & 79.1   \\
		{\sc KnowBert} embeddings $\rightarrow$ LR & \textbf{81.5}   \\
		\midrule
	    EntEmbeddings $\rightarrow$ Cosine  & 80.2   \\
		\bottomrule
	\end{tabular}
	\caption{Accuracy on the CAP test set. All baselines use logistic regression (LR) trained on the CAP training set. Ours predicts based on cosine similarity (no additional training required).}
	\label{tab:cap}
\end{table}

\paragraph{Classification Layer for Baselines} Following \citet{Mingda_Chen_19}, we train a simple classifier to make final predictions. Our feature vector of two entity representations $x_1$ and $x_2$ is a concatenation of $x_1$, $x_2$, element-wise product, and absolute difference: $[x_1, x_2, x_1 \odot x_2, |x_1 - x_2|]$. These are depicted in Figure~\ref{fig:downstream} as ``LR'' blocks.

This classifier is used for baselines only. Our approach only uses dot product or cosine similarity and does not require additional training. Since the size of our embeddings is generally larger, we avoid using a classifier in our setting because it would introduce more parameters than the baseline models, making fair comparison difficult.

\subsection{Embedding Model Hyperparameters}

We use pre-trained BERT-large uncased (24-layer, 1024-hidden, 16-heads, 340M parameters, whole word masking) \citep{Jacob_Devlin_19} for our mention and context encoder. All BERT hyperparameters are unchanged. The entity embedding matrix contains 10M (UFET type set) or 60M (Wiki type set) parameters.  We train our models with batch size 32 (8 $\times$ 4 gradient accumulation steps) using one NVIDIA V100 GPU for a week. We use the AdamW optimizer \citep{Kingma_14,Ilya_Loshchilov_18} with learning rate 2e-5 for BERT parameters and learning rate 1e-3 for the type embedding matrix. We use HuggingFace's Transformers library \citep{Thomas_Wolf_19} to implement our models.\footnote{Code and datasets used in our experiments are available at \url{https://github.com/yasumasaonoe/InterpretableEntityRepresentation}.}

\section{Results and Discussion}

\subsection{Coreference Arc Prediction (CAP)}

We compute our embeddings from an entity typing model trained on the UFET dataset \citep{Eunsol_Choi_18} for CAP (10k types). We choose this dataset because many of mention spans in the CAP examples are nominal expressions or pronouns, and the Wiki-Context dataset includes almost entirely mentions of proper nouns. To make a prediction if two mentions are coreferent, we compute $\text{sim}_{\text{cos}}(\mathbf{t}_1,\mathbf{t}_2)$ over the type vectors for each mention and check if this is greater than a threshold, which we set to 0.5.

Only our baselines use the CAP training set; our model does not train on this data. We compare our approach with the baselines described above as reported in \citet{Mingda_Chen_19}. Note that they use two different types of entity representations: one based on entity descriptions and another based on entity names only.   

Table~\ref{tab:cap} compares test accuracy on the CAP task. Although the {\sc KnowBERT} baseline achieves the highest accuracy, our entity representations reach comparable accuracy, 80.2, without training an additional classifier.  This validates our hypothesis that these embeddings are useful out-of-the-box and contain as much information as BERT-based embeddings, despite the constraints imposed by their explicit, interpretable structure.

\renewcommand{\arraystretch}{1}
\begin{table}[t]
	\centering
	\small
	\setlength{\tabcolsep}{4pt}
	\begin{tabular}{l c}
		\toprule
		\multicolumn{1}{l}{Model} & {Dev Acc.}\\
		\midrule
		{\sc BERT Base} $\rightarrow$ Cosine & 54.4  \\
		{\sc BERT Large} $\rightarrow$  Cosine & 52.7   \\
		Mention and Context Rep. $\rightarrow$ Cosine & 69.0  \\
		\midrule
	    EntEmbeddings $\rightarrow$ Cosine  & 80.1   \\
		\bottomrule
	\end{tabular}
	\caption{``Out-of-the-box'' accuracy on the CAP development set. We compare performance of BERT-base, BERT-large, and the mention and context representation of the embedding model with ours, using just cosine similarity and no classifier.}
	\label{tab:cap-ablation}
\end{table}

To further investigate the gains of our interpretable entity embeddings, we compare out-of-the-box performance (i.e., using cosine similarity instead of a task specific classifier) of BERT-base, BERT-large, and the hidden layer of the embedding model with ours. Table~\ref{tab:cap-ablation} shows development accuracy on CAP. BERT-base and BERT-large barely outperform random guessing (i.e., 50\%); we experimented with both the \textsc{[CLS]} and pooling methods and found pooling to work slightly better. We also compare to the mention and context representations $\mathbf{h}_{\texttt{[CLS]}}$ of our entity typing model. This latent representation is clearly better than the BERT representations but still underperforms our EntEmbeddings by 10 points.


\renewcommand{\arraystretch}{1}
\begin{table}[t]
	\centering
	\small
	\setlength{\tabcolsep}{4pt}
	\begin{tabular}{l c}
		\toprule
		\multicolumn{1}{l}{Model} & {Test Acc.}\\
		\midrule
		{\sc Most Frequent}  &  58.2 \\
		{\sc ELMo} Description   & 63.4 \\
		{\sc ELMo} Name    & 71.2 \\
		{\sc BERT Base} Description  & 64.7  \\
		{\sc BERT Base}  Name & 74.3  \\
		{\sc BERT Large}  Description  & 64.6 \\
		{\sc BERT Large}  Name & 74.8 \\
	    \midrule
	    EntEmbeddings $\rightarrow$ Cosine  & \textbf{84.8}  \\
		\bottomrule
	\end{tabular}
	\caption{Accuracy on the CoNLL-YAGO test set in the EntEval setting \cite{Mingda_Chen_19}. All baselines are from \citet{Mingda_Chen_19} and use logistic regression (LR) trained on the CoNLL-YAGO training set and the prior probability. Ours predicts based on cosine similarity (no additional training required).}
	\label{tab:conll}
\end{table}

\subsection{Named Entity Disambiguation (NED)}

We use the entity typing model trained on the Wiki-Context data (see Section~\ref{wiki-ctx}) to get the mention and context representation $\mathbf{t}$. In the CoNLL-YAGO setting, similar to past work \citep{Yasumasa_Onoe_19, Thibault_Fevry_20_b}, we prepend the document title and the first sentence to the input to enrich the context information. To obtain the candidate representations $\{\mathbf{c}_1, \mathbf{c}_2, ..., \mathbf{c}_j, ...\}$, we use the model trained on the Wiki-Description data, which is specialized for entity descriptions (see Section~\ref{wiki-desc}) similar to \citet{Dan_Gillick_19}. We choose Wikipedia datasets here because UFET does not support entity descriptions. We rank the candidate entities based on cosine similarity between $\mathbf{t}$ and $\mathbf{c}_j$, and the entity with the highest score is our model's prediction.

The {\sc Most Frequent} baseline chooses the most frequently observed entity for a given mention as a prediction, based on a prior probability $p_{\text{prior}}$ computed from link counts on Wikipedia.  All baselines except {\sc Most Frequent} combine the classifier output and the prior probability to make a prediction: $\argmax{c } \left[p_{\text{prior}}\left(c\right) + p_{\text{classifier}}\left(c\right)\right]$.\footnote{We adapt this technique from Chen et al. (2019) to be consistent with their setting.}

Table~\ref{tab:conll} lists test accuracy on the CoNLL-YAGO data. Our approach outperforms all baselines, indicating that our entity representations include useful information about entities out-of-the-box. Such a performance gap is expected since our entity representations can directly encode some factual knowledge from Wikipedia. However, these results also imply that pre-trained LMs \emph{do not} have enough factual information out-of-the-box; they may rely on in-domain fine-tuning to achieve high performance in the target domain, and often fail to generalize to new settings.

Note that while these accuracies are significantly below the supervised state-of-the-art (95\%), they are competitive with the ``zero-shot'' entity results from recent past work \cite{Nitish_Gupta_17,Yasumasa_Onoe_20}.

\renewcommand{\arraystretch}{1}
\begin{table}[t]
	\centering
	\small
	\setlength{\tabcolsep}{4pt}
	\begin{tabular}{l c c}
		\toprule
		\multicolumn{1}{l}{Model} & {Test Acc.}\\
		\midrule
		{\sc Most Frequent} &  64.6 \\
		{\sc ELMo} embeddings $\rightarrow$ LR + prior & 71.6  \\
		{\sc BERT Base} embeddings $\rightarrow$ LR + prior  & 69.6  \\
		{\sc BERT Large} embeddings $\rightarrow$ LR + prior  & 69.1 \\
		{\sc KnowBert} embeddings $\rightarrow$ LR + prior & 71.3   \\
	    \midrule
	    EntEmbeddings $\rightarrow$ Cosine  & \textbf{75.6} \\
		\bottomrule
	\end{tabular}
	\caption{Accuracy on the WLNED test set. All baselines use logistic regression (LR) trained on the WLNED training set and the prior probability. Ours predicts based on cosine similarity (no additional training required).}
	\vspace{-0.1cm}
	\label{tab:wlned}
\end{table}

Table~\ref{tab:wlned} shows test accuracy on the WLNED data. The general trend is similar to the CoNLL-YAGO results, and our approach outperforms all baselines. {\sc ELMo} embeddings achieve the highest accuracy, closely followed by {\sc KnowBERT} embeddings. 


\subsection{Reducing the Number of Types}

We show that our approach from Section~\ref{type-reduction} effectively prunes unnecessary types, and it leads to a compact task-specific entity typing model.

For the CAP dataset, we train a bilinear model with the dot scoring function and keep the top 1k types by their weights in $\mathbf{W}$ as the new type set. As can be seen in Table~\ref{tab:type-reduction}, the reduced  type set only results in a reduction of 1.2\% in development accuracy after removing 90\% of types.

To learn the type reduction in the CoNLL-YAGO setting, we convert the CoNLL-YAGO training data to a binary classification problem for simplicity by choosing positive and random negative entities. We train a model with the cosine scoring function and keep the top 5k types by weight as described in Section~\ref{type-reduction}. In Table~\ref{tab:type-reduction}, the reduced type set achieves the comparable development accuracy only using around 10\% of the original entity types. 

Combined, these results show that the computational tractability of our approach can be improved given a specific downstream task. While our large type vectors are domain-general, they can be specialized and made sparse for specific applications.

\renewcommand{\arraystretch}{1}
\begin{table}[t]
	\centering
	\small
	\setlength{\tabcolsep}{4pt}
	\begin{tabular}{l c c c c}
		\toprule
		\multicolumn{1}{l}{} & \multicolumn{1}{c}{} & \multicolumn{1}{c}{\#Types} & \multicolumn{1}{c}{} & \multicolumn{1}{c}{reduced} \\
		\cmidrule(r){3-3}  \cmidrule(r){5-5} 
		\multicolumn{1}{l}{Task} & \multicolumn{1}{c}{} & \multicolumn{1}{c}{Dev Acc.} & \multicolumn{1}{c}{} & \multicolumn{1}{c}{change} \\
		\midrule
		\multirow{2}{2.5cm}{CAP}
		&   & 10k $\longrightarrow$ 1k \: &  & 90\% \\
		&   & 80.1 $\longrightarrow$ 78.9 &  & $-1.2$ \\
		\addlinespace
		\multirow{2}{2.5cm}{CoNLL-YAGO}
		&   & 60k $\longrightarrow$ 5k \: &  & 92\% \\
		&   & 85.3 $\longrightarrow$ 85.0 &  & $-0.3$ \\

		\bottomrule  
	\end{tabular}
	\caption{Accuracy on the development sets before and after applying type reduction.}
	\label{tab:type-reduction}
	\vspace{-0.1cm}
\end{table}

\renewcommand{\arraystretch}{1}
\begin{table}[t]
	\centering
	\small
	\setlength{\tabcolsep}{4pt}
	\begin{tabular}{l c}
		\toprule
		\multicolumn{1}{l}{} & {Dev Acc.}\\
		\midrule
		EntEmbeddings $\rightarrow$ Cosine & 85.3    \\
		EntEmbeddings + \textbf{Debug} $\rightarrow$ Cosine & 87.0   \\
		\bottomrule 
	\end{tabular}
	\caption{Accuracy on the CoNLL-YAGO development set before and after applying debugging rules.}
	\label{tab:conll-debug}
	\vspace{-0.1cm}
\end{table}

\subsection{Debugging Model Outputs}

We investigate if simple rules crafted using domain knowledge can further fix errors as discussed in Section~\ref{Debuggability}. For CoNLL-YAGO, we create 11 rules and directly modify probabilities for certain types in entity representations $\mathbf{t}$. These rules are based on our observations of errors, in the same way that a user might want to inject domain knowledge while debugging their system. As described in Section~\ref{Debuggability}, this allows us to encode specific knowledge about domain entities (e.g., the particulars of championships for men's vs. women's tennis) that is unlikely to be encoded in our pre-trained model. The full set of rules is listed in Appendix A.


Table~\ref{tab:conll-debug} shows that by applying our 11 rules, which \emph{only} modify our type embeddings post-hoc, the development accuracy goes up by 1.7 points. Also note that such rule-based embedding changes are not task-specific, and change the embeddings for \emph{any} additional prediction task we wish to run on this data, whether it's entity linking, coreference, relation extraction, or more. Note that only a few tens of types are active and contribute to the final score substantially on any given example, so these can be handled with a relatively small set of rules. We believe that more generally, this could be a recipe for injecting knowledge when porting the system to new domains, as opposed to annotating training data.

\subsection{Analysis: Entity Typing Performance}

One important factor for our model is the performance of the underlying entity typing model. Table~\ref{tab:typing} shows the entity typing results on the development set of Wiki-Context, Wiki-Description, and UFET. On Wiki-Context, our entity typing model achieves 82.0 F1, which is fairly high given that there are 60,000 labels. All Wiki-Description development examples are unseen during the training time; thus, F1 is lower compared to Wiki-Context. The results on UFET are not directly comparable with past work\footnote{The SOTA performance on the original split is around 40 F1.} since we combine parts of the dev and test set in with the training set to have more training data. 

Overall, a BERT-based entity typing model can handle large number of entity types (10k or 60k) well. Some of the high performance here can be attributed to memorizing common entities in the training data.  However, we argue that this memorization is not necessarily a bad thing when the embeddings still generalize to work well on less frequent entities and in scenarios like CAP.

\renewcommand{\arraystretch}{1}
\begin{table}[h]
	\centering
	\small
	\setlength{\tabcolsep}{4pt}
	\begin{tabular}{l c c c c}
		\toprule
		\multicolumn{1}{l}{Model} & {\#Types} & {P} & {R} & {F1}\\
		\midrule
		{\sc Wiki-Context} & 60k & 86.7 & 77.7 & 82.0 \\
		{\sc Wiki-Description} &60k & 77.6 & 71.2 & 74.2 \\
		{\sc UFET} & 10k & 54.6 & 40.5 & 46.5 \\
		\bottomrule
	\end{tabular}
	\caption{Macro-averaged P/R/F1 on the development sets.}
	\label{tab:typing}
\end{table}

\section{Related Work}

Some past work learns static vectors for millions of predefined entities. \newcite{Ikuya_Yamada_16} and \newcite{Yotam_Eshel_17} embed words and entities in the same continuous space particularly for NED. \citet{Jeffrey_Ling_20} learn general purpose entity embeddings from context and entity relationships in a knowledge base while \citet{Thibault_Fevry_20} does not rely on that structured information about entities. Our approach only stores type embeddings which can be substantially smaller than the entity embedding matrix.

Entity typing information has been used across a range of NLP tasks, including models for entity linking and coreference \citep{Durrett_Klein_14}. In entity linking specifically, typing has been explored for cross-domain entity linking \citep{Nitish_Gupta_17,Yasumasa_Onoe_20}. Past work by \citet{Jonathan_Raiman_18} has also explored learning a type system for this task. Our approach to learning types starts from a large set and filters it down, which is a simpler problem. A range of approaches have also considered augmenting pre-trained models with type information \cite{Matthew_Peters_19}; however, in these models, the types inform dense embeddings which are still uninterpretable.

A related thrust of the literature has looked at understanding entities using interpretable embeddings based around feature norms \citep{Ken_McRae_05}; this has advantages for learning in few-shot setups \citep{Su_17}. However, most of this past work has used embeddings that are much lower-dimensional than ours, and don't necessarily to scale to broad-domain text or all of Wikipedia.

Another line of past work tests if type information or other knowledge is captured by pre-trained LMs. \citet{Matthew_Peters_18} report that ELMo performs well on word sense disambiguation and POS tagging. Some other work also investigates models' ability to induce syntactic information by measuring accuracy of a probe \citep{Kelly_Zhang_18, John_Hewitt_19_a, John_Hewitt_19_b}. However, there is significant uncertainty about how to calibrate such probing results \citep{Elena_Voita_20}; our model's representations are more directly interpretable and don't require post-hoc probing.

Lastly, our work is distinct from SPINE \citep{Anant_Subramanian_07}, a past technique for learning sparse interpretable embeddings. However, this technique requires an additional step to reveal their interpretability. Each dimension of our entity representations has a name (i.e., a fine-grained type) with a probability, and thus it is immediately interpretable.

\section{Conclusion} In this work, we presented an approach to creating interpretable entity representations that are human readable and achieve high performance on entity-related tasks out of the box. We show that it is possible to reduce the size of our type set in a learning-based way for particular domains. In addition, these embeddings can be post-hoc modified through simple rules to incorporate domain knowledge and improve performance.

\section*{Acknowledgments}

Thanks to the anonymous reviewers for their helpful comments, members of the UT TAUR lab for helpful discussion, Pengxiang Cheng for constructive suggestions, and Mingda Chen for providing the details of experiments. This work was partially supported by NSF Grant IIS-1814522, NSF Grant SHF-1762299, a gift from Arm, and an equipment grant from NVIDIA. The authors acknowledge the Texas Advanced Computing Center (TACC) at The University of Texas at Austin for providing HPC resources used to conduct this research.

This material is also based on research that is in part supported by the Air Force Research Laboratory (AFRL), DARPA, for the KAIROS program under agreement number FA8750-19-2-1003. The U.S. Government is authorized to reproduce and distribute reprints for Governmental purposes notwithstanding any copyright notation thereon. The views and conclusions contained herein are those of the authors and should not be interpreted as necessarily representing the official policies or endorsements, either expressed or implied, of the Air Force Research Laboratory (AFRL), DARPA, or the U.S. Government.

\bibliography{emnlp2020}
\bibliographystyle{acl_natbib}

\section*{Appendix A: Debugging Rules}\label{app:rules}

(See Table~\ref{tab:rules})

\renewcommand{\arraystretch}{1}
\begin{table*}[h]
	\centering
	\small
	\begin{tabular}{p{4cm} p{5cm} p{5cm}}
		\toprule
		\multicolumn{1}{l}{Rule}  & {Types to be set to 1} & {Types to be set to 0}\\
		\midrule
		\emph{fed cup} is in the context.  & \texttt{women's, tennis, teams, sports} & \texttt{davis cup teams, davis cup} \\
		\midrule
		\emph{soccer} is in the context.  & \texttt{football} & \texttt{uefa member associations} \\
		\midrule
		\emph{cricket} is in the context.  & \texttt{england in international cricket, men's, national cricket teams, in english cricket} & \texttt{women's, women, women's national cricket teams,  football} \\
        \midrule
        \emph{tennis} is in the context, and  the mention is \emph{washington}. & \texttt{tennis, living people} & \texttt{cities, in washington (state), in washington, d.c, established, establishments,
            capital districts and territories, populated, 'places} \\
        \midrule
        The mention is \emph{wall street}. &  \texttt{exchanges, stock} &  \texttt{streets, tourist}\\
        \midrule
        \emph{soccer} and \emph{1996} are in the context, and the mention is \emph{worldcup}. &  \texttt{1998} & \texttt{1996} \\
        \midrule
        \emph{baseball} and \emph{new york} are in the context, and the mention is \emph{chicago}. &  \texttt{chicago white sox} & \texttt{chicago cubs}\\
        \midrule
        \emph{yeltsin} is in the context, and the mention is \emph{lebed}. &  \texttt{living people, of russia} & \\
        \midrule
        \emph{venice festival} is in the context, and the mention is \emph{jordan}. &  \texttt{living people, people, irish, irish male novelists, 1950 births, male screenwriters,
           bafta winners (people), writers, for best director winners,
           people from dublin (city), 20th-century irish novelists} & \texttt{member states of the organisation of islamic cooperation, of the organisation of islamic cooperation, in jordan, territories, countries, states, of the arab league, member, western asian countries, member states of the arab league, member states of the united nations, jordan, tourism}\\
        \midrule
        \emph{baseball} is in the context.  & \texttt{major, baseball} & \texttt{soccer, football, major league soccer, professional sports leagues in canada, professional, in the united states, in canada} \\
        \midrule
        \emph{squash} is in the context, and the mention is \emph{jansher}. &  \texttt{1969 births} & \texttt{1963 births}\\
		\bottomrule
	\end{tabular}
	\caption{Debugging rules applied for the CoNLL-YAGO development set.}
	\label{tab:rules}
\end{table*}

\end{document}